\documentclass[]{bytedance_seed}

\usepackage[toc,page,header]{appendix}

\usepackage{natbib}

\usepackage{CJKutf8}

\usepackage{xargs}  

\usepackage{todonotes}  

\usepackage{multirow}

\usepackage{cleveref}

\usepackage{amsmath}
\usepackage{dsfont}

\usepackage{subcaption}

\usepackage{svg}

\usepackage{mathrsfs}
\usepackage{adjustbox}
\usepackage{amsmath}
\usepackage{multirow}
\usepackage{multirow}
\usepackage{multicol}
\usepackage{tcolorbox}
\usepackage{changepage}
\usepackage{graphicx}
\usepackage{amssymb}
\usepackage{array}
\usepackage{bm}

\usepackage{minitoc}

\title{Seedream 4.0: Toward Next-generation Multimodal Image Generation}

\author[]{ByteDance Seed}

\abstract{
We introduce Seedream 4.0, an efficient and high-performance multimodal image generation system that unifies text-to-image (T2I) synthesis, image editing, and multi-image composition within a single framework. 
We develop a highly efficient diffusion transformer with a powerful VAE which also can reduce the number of image tokens considerably. This allows for efficient training of our model, and enables it to fast generate native high-resolution images (e.g., 1K-4K). Seedream 4.0 is pretrained on billions of text–image pairs spanning diverse taxonomies and knowledge-centric concepts. Comprehensive data collection across hundreds of vertical scenarios, coupled with optimized strategies, ensures stable and large-scale training, with strong generalization. By incorporating a carefully fine-tuned VLM model, we perform multi-modal post-training for training both T2I and image editing tasks jointly. For inference acceleration, we integrate adversarial distillation, distribution matching, and quantization, as well as speculative decoding. It achieves an inference time of up to 1.4 seconds for generating a 2K image (without a LLM/VLM as PE model). Comprehensive evaluations reveal that Seedream 4.0 can achieve state-of-the-art results on both T2I and multimodal image editing. In particular, it demonstrates exceptional multimodal capabilities in complex tasks, including precise image editing and in-context reasoning, and also allows for multi-image reference, and can generate multiple output images. This extends traditional T2I systems into an more interactive and multidimensional creative tool, pushing the boundary of generative AI for both creativity and professional applications.
We further scale our model and data as Seedream 4.5, which consistently outperforms Seedream 4.0 in both T2I and image editing tasks.
Seedream 4.0 and Seedream 4.5 are accessible on \href{https://www.volcengine.com/experience/ark?launch=seedream}{Volcano Engine}\textsuperscript{$\alpha$}.
}

\checkdata[\textsuperscript{$\alpha$}Model ID]{Doubao-Seedream-4.0, \url{https://seed.bytedance.com/seedream4_0}} 
\checkdata[\textsuperscript{$\alpha$}Model ID]{Doubao-Seedream-4.5, \url{https://seed.bytedance.com/zh/seedream4_5}}

\begin{document}
\begin{CJK*}{UTF8}{gbsn}

\maketitle

\definecolor{chinese_red}{HTML}{8B4513}
\definecolor{english_blue}{HTML}{4169E1}

\begin{figure}[ph]
\begin{center}
\vspace{-30pt}
\includegraphics[height=5cm]{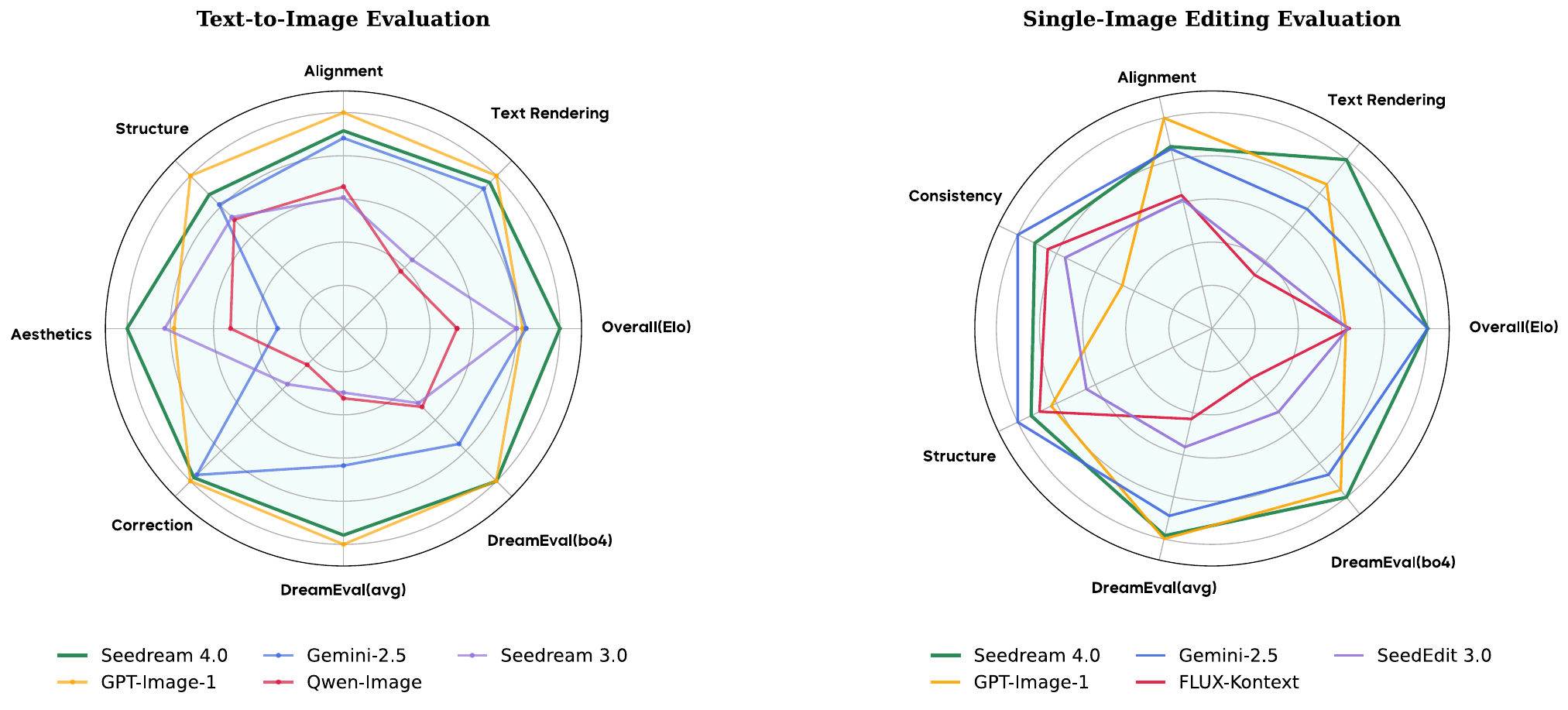}
\end{center}
\label{fig:overall_eval}
\vspace{-5pt}
\caption{Overall evaluation. Left: Text-to-Image results; Right: Image-Editing results.
The Elo scores are obtained from the Artificial Analysis Arena. Seedream 4.0 ranks first in both T2I and image-editing leaderboards, by 09/18/2025.
}
\vspace{-10pt}
\end{figure}

\begin{figure}[pt]
\begin{center}
\includegraphics[width=0.88\linewidth]{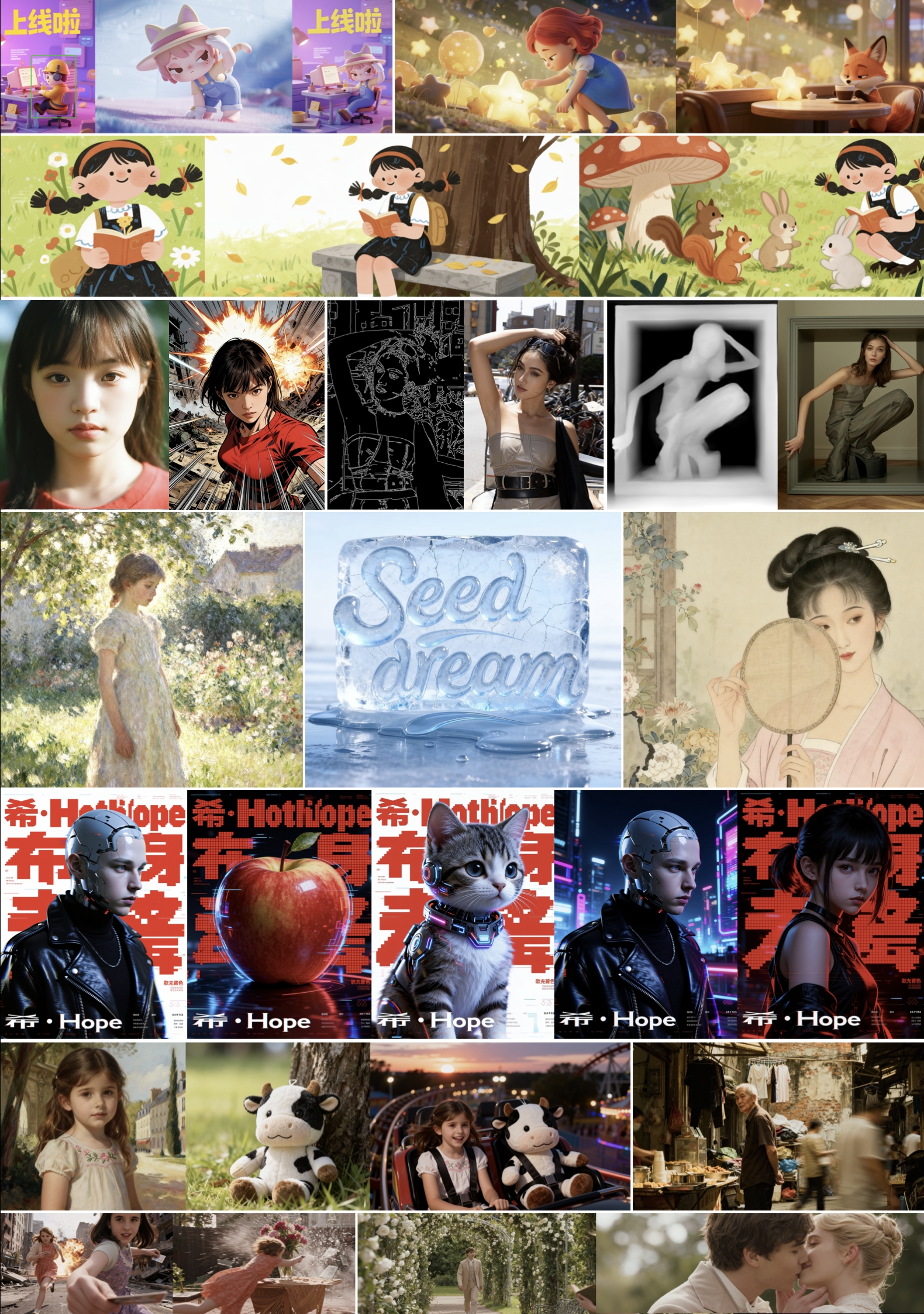}
\end{center}
\label{fig:teaser}
\vspace{-1pt}
\caption{Seedream 4.0 visualization.}
\end{figure}


\clearpage

\tableofcontents

\newpage

\section{Introduction}

Diffusion models have ushered in a new era in generative AI, enabling the synthesis of images with remarkable fidelity and diversity. Building on recent advances in diffusion transformers (DiTs), state-of-the-art open-source and commercial systems have emerged, such as Stable Diffusion \cite{rombach2022high}, FLUX series \cite{flux2023,labs2025flux1kontextflowmatching}, Seedream models \cite{gong2025seedream,gao2025seedream,shi2024seededit}, GPT-4o image generation \cite{gpt-4o} and Gemini 2.5 flash \cite{gemini2.5}. However, as the demand for higher image quality, greater controllability, and strong multimodal capabilities (e.g., text-to-image (T2I) synthesis and image editing) increases, current models often have a critical scalability bottleneck. 

In this paper, we introduce Seedream 4.0, a powerful multimodal generative model engineered for scalability and efficiency. We develop an efficient and scalable DiT backbone, which substantially increases the model capacity while reducing the training and inference FLOPs considerably. To further enhance model efficiency, especially for high-resolution image generation, we have developed an efficient Variational Autoencoder (VAE) with a high compression ratio, significantly reducing the number of image tokens in latent space.  This architectural design (including our DiT and VAE) makes our model highly efficient, easily scalable, and hardware-friendly in both training and inference. Our training strategy is meticulously designed to unlock the full potential of our architecture, achieving more than 10$\times$ inference acceleration compared to Seedream 3.0~\cite{gao2025seedream}, while having significantly better performance.
%
This allows the model to be trained effectively on billions of text–image pairs at native image resolutions ranging from 1K to 4K, covering a wide range of taxonomy and knowledge-centric concepts. 

In the post-training stage, we incorporate a VLM model with a strong understanding of multimodal inputs into our system， as designed in SeedEdit \cite{shi2024seededit}, which enables it with strong multimodal generation ability. We pioneer a joint post-training that integrates both T2I generation and image editing through a causal diffusion designed in the DiT framework. 
Our post-training stage starts with Continuing Training (CT) to broaden the model's foundational knowledge and multi-task proficiency. This is followed by Supervised Fine-Tuning (SFT), which works to inculcate specific artistic qualities. Subsequently, we implement Reinforcement Learning from Human Feedback (RLHF) to meticulously align the model's outputs with nuanced human preferences. Then a Prompt Engineering (PE) module is developed to unlock the full potential of the model across a diverse spectrum of user inputs.
To achieve ultra-fast inference on our DiT, we propose a holistic acceleration system centered on an adversarial learning framework. This core algorithmic advance is synergistically combined with hardware-aware quantization and speculative decoding, culminating in a system that delivers a second-level generation experience without quality degradation.

In summary, Seedream 4.0 presents the following advantages:
\begin{itemize}
    \item \textbf{Efficient and Scalable Architecture.} Our model contains a carefully designed DiT architecture together with a powerful VAE, yet having a high compression ratio. This results in a highly efficient architecture that achieves more than 10$\times$ training and inference acceleration (measured by compute flops) compared to Seedream 3.0 \cite {gao2025seedream}, with significantly better performance obtained.
     %
    The efficient architecture of Seedream 4.0 allows for strong scalability in terms of modal capacity, task coverage, and multimodel generalization.
    \item \textbf{Strong Multi-modal Generation.} We extended the Seededit 3.0 \cite{seededit2024} architecture for multimodal generation, and perform multimodal joint post-training of T2I and image editing tasks on our pre-trained DiT model. This enables Seedream 4.0 with strong multi-modal capability that allows for single- or multi-image inputs and outputs.
    \item \textbf{Professional creation scenarios.}  Beyond artistic imagery, Seedream 4.0 exhibits a strong capability to generate structured, professional and knowledge-based content, such as charts, formulas, and design materials, bridging the gap between creative generation and practical industry applications.
    %
    \item \textbf{Ultra-fast inference speed.} With efficient architecture design, we further optimize our framework aggressively to achieve extreme inference acceleration. This allows our model to perform ultrafast image generation and editing at high resolutions (e.g., 2K or 4K), greatly enhancing user interaction experience and production efficiency.
\end{itemize}
Seedream 4.0 has been successfully integrated into multiple platforms, including Doubao and Jimeng as of September 2025.
It can also be accessible on \href{https://www.volcengine.com/experience/ark?launch=seedream}{Volcano Engine}\textsuperscript{$\alpha$}.
We believe Seedream 4.0 will become a practical tool to improve productivity in all aspects of work and daily life.
\footnotetext[1]{https://www.doubao.com/chat/create-image}
\footnotetext[2]{https://jimeng.jianying.com/ai-tool/image/generate}


\section{Data, Model Training and Acceleration}

\subsection{Model Pre-training}

\textbf{Data.} In Seedream 3.0, we introduced a dual-axis collaborative data sampling framework that jointly optimizes pre-training data along two dimensions: visual morphology and semantic distribution. However, we observed two limitations when applying a purely top-down resampling strategy:
\begin{itemize}
\item It disproportionately favors natural images in the overall distribution.
\item It underrepresents fine-grained, knowledge-centric concepts (e.g., instructional content and mathematical expressions).
\end{itemize}
To address these issues, we redesigned the pipeline specifically for knowledge-related data, including instructional images and formulae.

In our pipeline, knowledge data are categorized into natural and synthetic subsets. For natural images, we collect high-quality figures from PDF documents that span in-house textbooks, research articles, and novels. We first deploy a low-quality classifier to filter out undesirable samples (e.g. blurred images, cluttered, or noisy backgrounds). Next, we train a difficulty rating classifier with three levels: easy, medium, and hard, and we annotate all images accordingly. Images with extremely difficulty are down-sampled during pre-training. For synthetic data, we used both OCR output and LaTeX source code~(when available) to generate diverse formula images that vary in structure (layout, symbol density) and resolution. This synthesis strategy broadens the coverage of fine-grained concepts and mitigates the biases introduced by top-down resampling.

Beyond knowledge-related data, we introduce several module-level upgrades compared to our previous version. (1) we train a text-quality classifier to detect low-quality text in original caption; (2) we combine semantic and low-level visual embeddings in the deduplication pipeline to boost the deduplication results, balancing fine-grained distribution; (3) we refine the captioning model for finer-grained visual descriptions; and (4) we adopt a stronger cross-modal embedding for image–text alignment, substantially improving our multimodal retrieval engine.

\textbf{Training Strategies.} Similar to Seedream 3.0~\cite{gao2025seedream}, we adopt multi-stage training to improve training efficiency. In the first stage, we train our DiT at an average resolution of 512$^2$ (with different aspect ratios). In the second stage, we fine-tune our model at higher resolutions spanning from $1024^2$ to $4096^2$. Thanks to the efficient design of our model, training on these higher resolutions up to 4K is still effective.

\textbf{Training Infrastructures.} To enable efficient large-scale pre-training of the DiT model, we design a highly optimized training system that emphasizes hardware efficiency, scalability, and robustness. The key components are summarized as follows.

\textit{Parallelism and Memory Optimization.} We employ Hybrid Sharded Data Parallelism (HSDP) to efficiently shard weights and support large-scale training without resorting to tensor or expert parallelism. Memory usage is optimized through timely release of hidden states, activation offloading, and enhanced FSDP support, enabling training of large models within available GPU resources.

\textit{Kernel and Workload Optimization.} Performance-critical operations are accelerated by combining torch.compile with handcrafted CUDA kernels and operator fusion, reducing redundant memory access. To address workload imbalance from variable sequence lengths, we introduce a global greedy sample allocation strategy with asynchronous pipelines, achieving more balanced per-GPU utilization.

\textit{Fault Tolerance.}
Multi-level fault tolerance is built into the system, including periodic checkpointing of model, optimizer, and dataloader states, pre-launch health checks to exclude faulty nodes, and reduced initialization overhead. These measures ensure stability and sustained throughput during long-term distributed training.

\subsection{Model Post-training}
We perform an intensive post-training to enhance the multimodal capabilities of our model, including T2I, single image editing, and multi-image reference and output.
Specifically, we perform joint training of multiple tasks through multistage post-training, containing continuing training (CT), supervised fine-tuning (SFT), and human feedback alignment (RLHF) \cite{xu2024imagereward,wu2025rewarddance,xue2025dancegrpo,liu2025flow}; and a carefully fine-tuned prompt engineering (PE) model was also developed. 
Performance is improved consistently and significantly at each substage, resulting in higher performance compared to models trained on individual tasks. In particular, the CT stage mainly enhances the instruct following ability for image editing, and the SFT further improves the consistency between the reference and edited images considerably. 

We construct a large amount of editing data that is used in the CT and SFT stages. Each data sample typically has a reference image and a target image, with an editing instruction.  
Image captions are produced for both reference and target images. We designed three types of caption with different levels of detail, which function as a form of data augmentation during training. In addition, we encourage the use of consistent terminology in captions to describe similarities between reference and target images. 


We trained an end-to-end vision language model (VLM) as our PE model based on Seed1.5-VL~\cite{guo2025seed1}. This VLM model processes user input, including a text prompt, a single reference image, or multiple images, and generates the corresponding outputs (for example, the captions of the reference image and the target or predicted image, as did SeedEdit 3.0 \cite{seededit2024}), which are then used as input to the DiT model. The functions of the PE model also include task routing, prompt rewriting (with auto-thinking), and optimal aspect ratio estimation. To balance latency and performance, our model dynamically adjusts its thinking budgets based on task complexity, inspired by AdaCoT~\cite{lou2025adacot}. This integrated approach enables Seedream 4.0 to better address user intentions, perform complex reasoning, and generate a series of images from a single request.

\subsection{Model Acceleration}

\textbf{Efficient, High-Quality Synthesis.}
Our acceleration framework integrates principles from Hyper-SD~\cite{ren2025hyper}, RayFlow~\cite{shao2025rayflow}, APT~\cite{lin2025diffusion}, and ADM~\cite{lu2025adversarial} to accelerate Diffusion Transformers (DiTs). Our approach establishes an innovative paradigm where each sample follows an optimized, adaptive trajectory, rather than a shared path to a Gaussian prior. This customization minimizes trajectory overlap and reduces instability. To learn these paths effectively, we employ an adversarial matching framework that replaces fixed divergence metrics, circumventing mode collapse and significantly improving generation stability and sample diversity. This is achieved through a two-stage process, starting with a robust Adversarial Distillation post-training (ADP) stage that uses a hybrid discriminator to ensure a stable initialization. Following this, an Adversarial Distribution Matching (ADM) framework employs a learnable, diffusion-based discriminator for fine-tuning, enabling a more fine-grained matching of complex distributions. Our unified pipeline enables highly efficient few-step sampling, drastically reducing the Number of Function Evaluations (NFE) while achieving results that match or surpass baselines requiring dozens of steps across key dimensions like aesthetic quality, text-image alignment, and structural fidelity, effectively balancing quality, efficiency, and diversity.

\textbf{Quantization.} 
To further boost inference performance without quality loss, we employ a hardware-aware framework combining quantization and sparsity. Our approach uses an adaptive 4/8-bit hybrid quantization, which involves offline smoothing to handle outliers, a search-based optimization to find the best granularity and scaling for sensitive layers, and post-training quantization (PTQ) to finalize parameters. We co-design this with efficient, hardware-specific operators for various bit widths and granularities to maximize performance.

\textbf{Speculative Decoding for PE.}
Our method builds upon the foundational work of Hyper-Bagel\cite{lu2025hyperbagelunifiedaccelerationframework} to address the inherent uncertainty in speculative decoding that arises from stochastic token sampling. Our solution conditions feature prediction on both the preceding feature sequence and a token sequence advanced by one timestep. This provides a deterministic target that resolves sampling ambiguity and significantly enhances prediction accuracy. We further improve this process by incorporating a loss function on Key-Value (KV) caches to enable efficient reuse during inference and an auxiliary cross-entropy loss on logits to refine the draft model.

\section{Model Performance}

\begin{figure}[t]
    \centering
    \includegraphics[width=\linewidth]{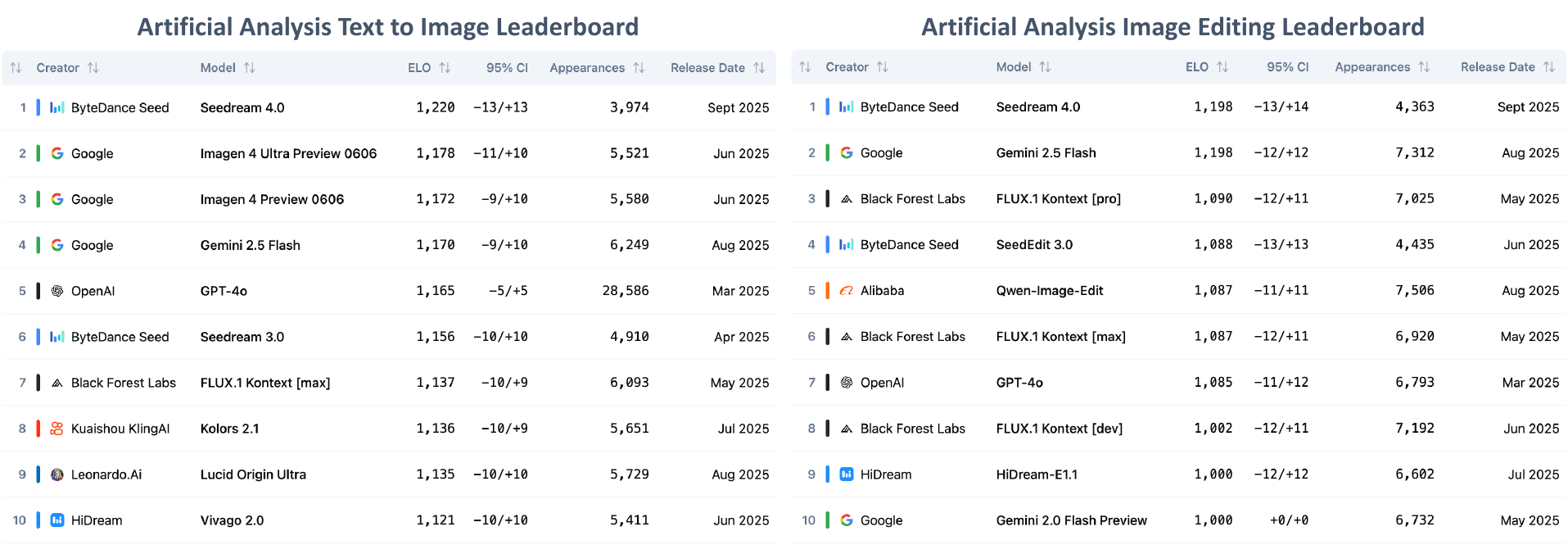}
    \caption{Results from the Artificial Analysis Arena\protect\footnotemark[1]: Seedream 4.0 leads in both the Text-to-Image and Image Editing tracks.}
    \label{fig:aa_seedream4.0}
\end{figure}

In this section, we present a comprehensive evaluation of Seedream 4.0. 
First, we report the results of an overall ELO score from a public platform, Artificial Analysis Arena~\cite{aa_elo}, as shown in Figure~\ref{fig:aa_seedream4.0}.
By maintaining a real-time competitive arena, Artificial Analysis Arena continuously incorporates newly released models and provides dynamic leaderboards.
The participants cover the Seedream series (Seedream 3.0~\cite{gao2025seedream}, Seedream 4.0, and SeedEdit 3.0~\cite{seededit2024}) alongside other leading models, including GPT-Image-1~\cite{openai2024gpt4ocard}, Gemini-2.5 Flash Image~\cite{gemini2.5} (abbreviated as Gemini-2.5), and open-source models such as Qwen-Image~\cite{wu2025qwenimagetechnicalreport}, and FLUX-Kontext\cite{labs2025flux1kontextflowmatching}.
The results indicate that Seedream 4.0 ranks first in both the single-image editing and text-to-image tracks.

To further explore the fine-grained capabilities of Seedream 4.0, we provide both human evaluation and automated benchmarking results. Across various tasks and dimensions, Seedream 4.0 consistently delivers top-tier performance.
Finally, we highlight its excellent multimodal image generation capabilities. By flexibly combining input and output modalities, Seedream 4.0 can support a wide range of creative applications. We also present illustrative examples of these functionalities, while noting that further possibilities remain to be explored through user interaction.

\footnotetext[1]{
Data as of 17:00 (Beijing Time), September 18.
\url{https://artificialanalysis.ai/text-to-image/arena?tab=Leaderboard}}

\subsection{Comprehensive Human Evaluation}
To benchmark the performance of Seedream 4.0 against other top-tier image generation models, we constructed a comprehensive multimodal benchmark, MagicBench 4.0. The benchmark covers three major task categories: text-to-image (T2I) generation, single-image editing, and multiimage editing. The three tracks consist of 325 prompts, 300 prompts, and 100 prompts, respectively; and each prompt is provided in both Chinese and English. 
In the following sections, we present a detailed analysis of various models in these tasks.


\subsubsection{Text-to-Image}

\begin{figure*}[h]
\centering
\includegraphics[width=\linewidth]{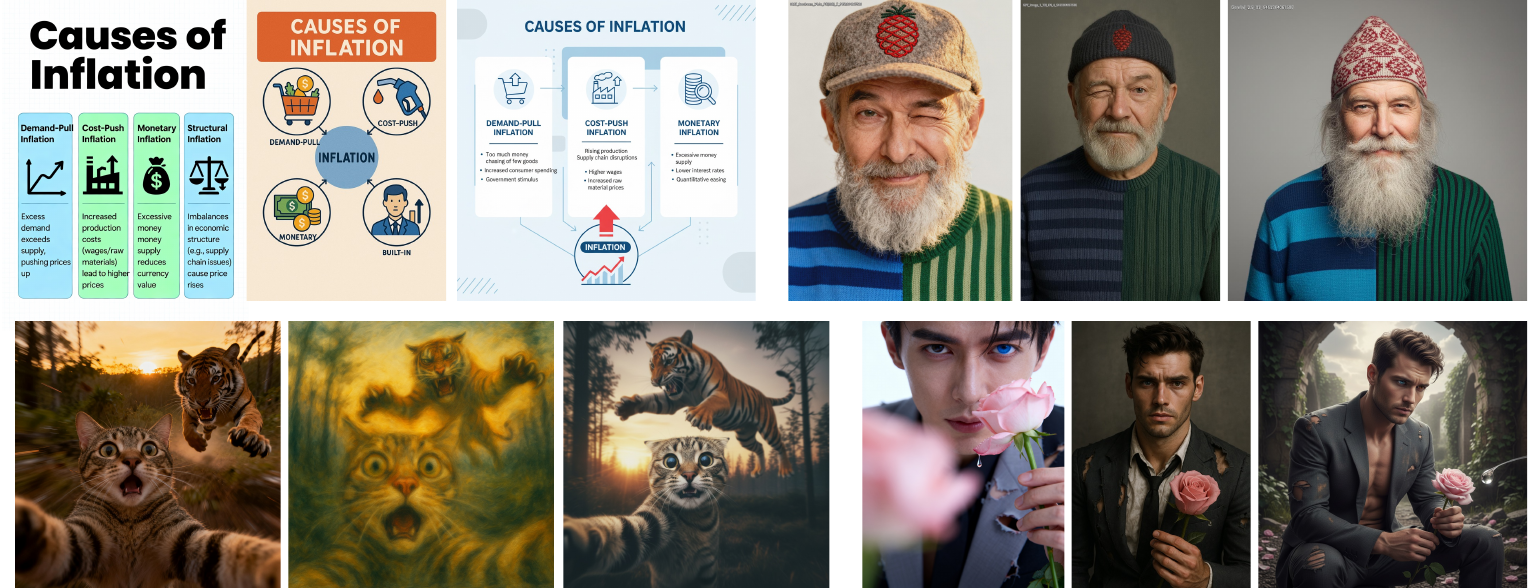}
\caption{Text-to-Image comparisons. Within each group, images are arranged from left to right as Seedream 4.0, GPT-Image-1, and Gemini-2.5. \textbf{Prompt1}: Draw an infographic showing the \textcolor{red}{causes of inflation}.; \textbf{Prompt2}: A studio-style \textcolor{red}{portrait of an elderly man: he ..., with his \textcolor{red}{left eye tightly closed in a witty and amusing manner}.}. \textbf{Prompt3}: \textcolor{red}{狸花猫第一视角自拍，老虎在后面追}，超现实主义，写实，黄昏背景; \textbf{Prompt4}: A handsome male model whose \textcolor{red}{left eye pupil is blue} wears a torn suit with a pink rose in his \textcolor{red}{left hand}; there is only \textcolor{red}{a single dewdrop} clearly visible on the \textcolor{red}{edge of the rose petal}, as if it were \textcolor{red}{about to fall} from the petal to the ground in the next second.}
\label{fig:t2i}
\end{figure*}

As a fundamental capability of image generation models, T2I generation has always been a key focus of the Seedream series. In addition to conventional dimensions such as prompt alignment, structural stability, and visual aesthetics, we evaluate the model's performance in two additional aspects: dense text rendering and content understanding. The latter is particularly relevant to prompts that require advanced in-context reasoning or specialized domain knowledge. As shown in Figure 1, Seedream 4.0 demonstrates significant improvements in all evaluation dimensions compared to its predecessor. In particular, it substantially outperforms competing models in visual aesthetics. Figure~\ref{fig:t2i} presents several challenging cases of T2I by comparing Seedream 4.0 with GPT-Image-1 and Gemini-2.5. These samples illustrate the model's capability in ensuring strong instruction adherence, precise text rendering, and stable visual quality across models, and Seedream 4.0 stands out for its superior visual impact, including a dynamic sense of motion, natural lighting, and coherent color composition.

\subsubsection{Single-Image Editing}

\begin{figure*}[h]
\centering
\includegraphics[width=\linewidth]{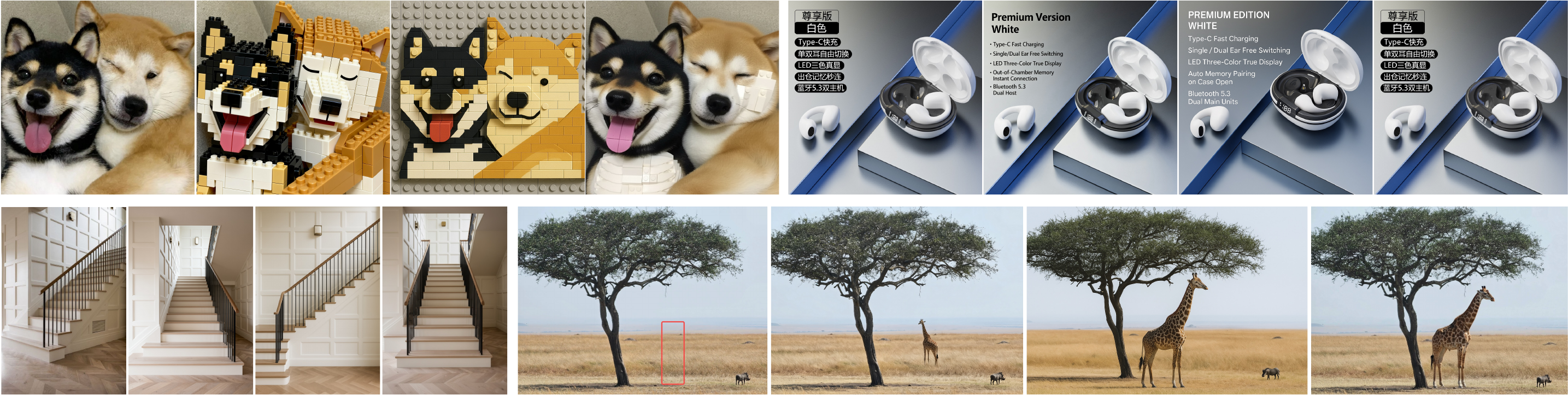}
\caption{Single-Image Editing Comparisons. Within each group, images are arranged from left to right as the original image, Seedream 4.0, GPT-Image-1, and Gemini-2.5. \textbf{Prompt1}: 把图中的两只狗改为\textcolor{red}{乐高风格}.; \textbf{Prompt2}: Translate the text of the product image into English. \textbf{Prompt3}: Please modify the photo's perspective to make it look like it was taken \textcolor{red}{directly facing the stairs}.; \textbf{Prompt4}: 在图中\textcolor{red}{红框位置}添加一只长颈鹿.}
\label{fig:single_editing}
\end{figure*}

Seedream 4.0 integrates both editing and generation capabilities in a unified pipeline, enabling them to mutually enhance each other and delivering performance that exceeds the previous version, SeedEdit 3.0. A crucial challenge in image editing lies in the trade-off between instruction following and consistency, which is also what our evaluation focuses on. In addition, we also consider structural integrity and text-editing performance. As shown in Figure 1, the results reveal distinct patterns in all leading models. GPT-Image-1 achieves the highest
accuracy in instruction following, but ranks lowest in consistency, which is a limitation that has been widely noted.
In contrast, Gemini-2.5 excels at preservation, but shows limited capability in instruction following, particularly for tasks such as style transfer and viewpoint transformation, as shown in Figure~\ref{fig:single_editing}; it also struggles with text editing, especially in Chinese. Seedream 4.0, by comparison, demonstrates more balanced performance in all dimensions. It supports a wide range of editing tasks, maintains strong consistency, and thus achieves a
higher level of practical usability.

\subsubsection{Multi-Image Editing}

\begin{figure*}[h]
\centering
\includegraphics[width=\linewidth]{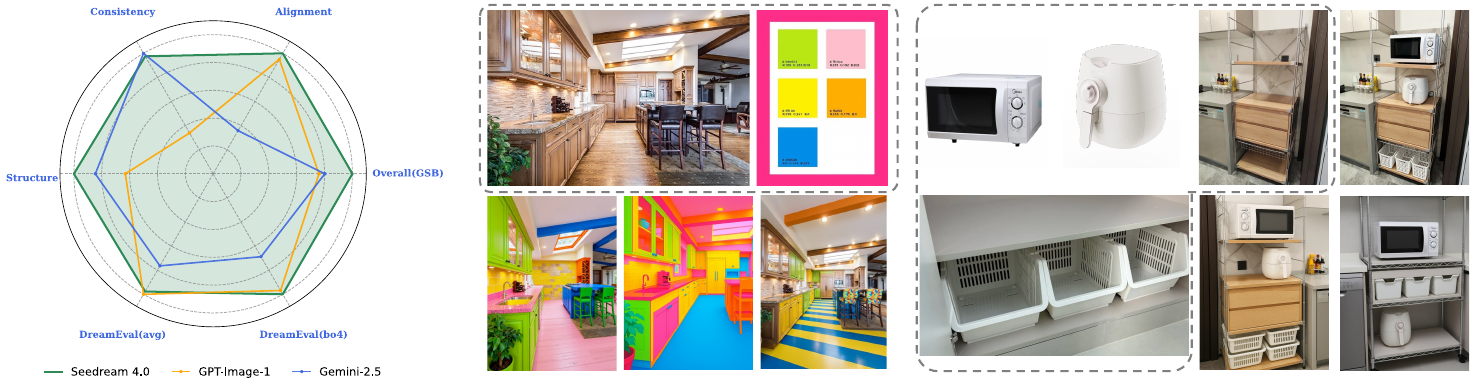}
\caption{Multi-image editing comparison. Left: overall evaluation results. Right: two examples, with dashed boxes indicating input references and outputs shown left-to-right as Seedream 4.0, GPT-Image-1, and Gemini-2.5. In the rightmost example, the top-right image is from Seedream 4.0. \textbf{Prompt1}: \textcolor{red}{Referring to the color chart} in picture two, transform the room and all the furniture into a dopamine style.; \textbf{Prompt2}: Place the items in \textcolor{red}{image one on the top shelf} of the cabinet. Place the items in \textcolor{red}{image two on the middle shelf}. Place the items in \textcolor{red}{image three on the bottom shelf} of the cabinet.}
\label{fig:multi_editing}
\end{figure*}

Multi-image editing goes beyond the simple combination of multiple images; it requires models to perform rich in-context understanding of objects across different inputs. We compare the performance of Seedream 4.0 against GPT-Image-1 and Gemini-2.5 using an overall metric (GSB) with three objective dimensions: alignment of the instruction, consistency and structure. As illustrated in Figure~\ref{fig:multi_editing}, the results mirror those of single-image editing: GPT-Image-1 shows strength in instruction responsiveness but weak consistency, while Gemini-2.5 excels in preservation, but falls short in responsiveness. In contrast, Seedream 4.0 performs at the highest levels in all dimensions, outperforming the other two models by almost 20\% in the GSB metric. A crucial consideration in multi-image editing is structural integrity. We observed that as the number of reference images increases, the outputs of other models tend to suffer from structural degradation. However, Seedream 4.0 maintains more stable and coherent structures, demonstrating robust performance even when provided with more than ten reference images.

\subsection{Automatic Evaluation with DreamEval}

\begin{figure*}[h]
\centering
\includegraphics[width=\linewidth]{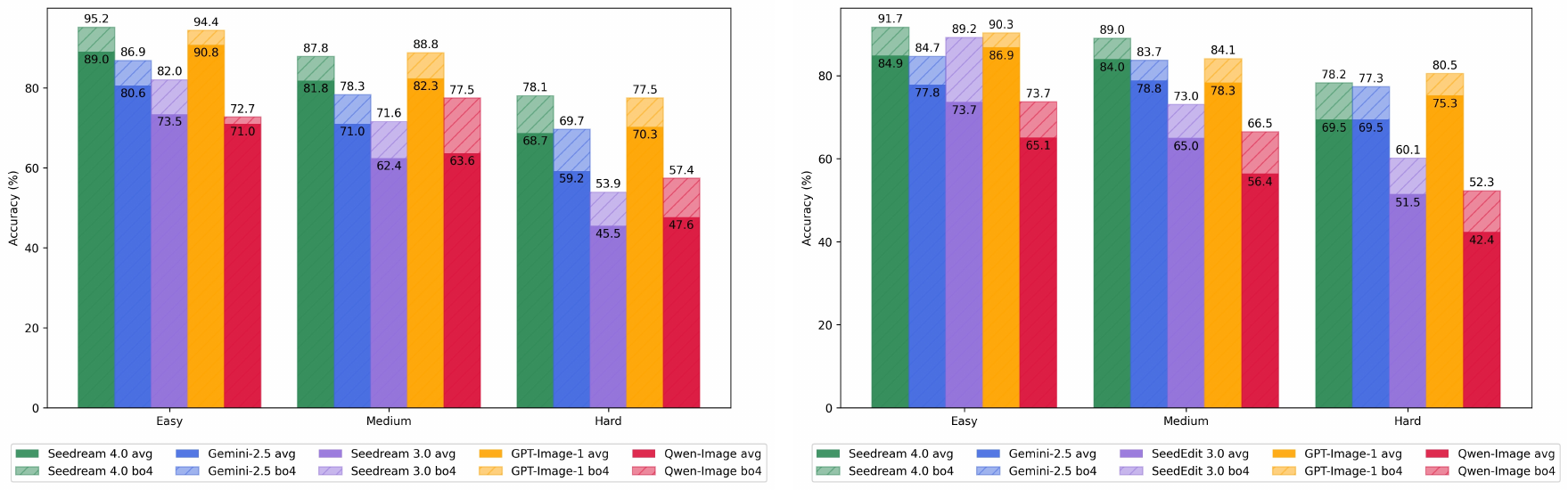}
\caption{Automatic Evaluation with DreamEval. Left: Text-to-Image; Right: Single-Image Editing.}
\label{fig:dreameval}
\end{figure*}

Automatic evaluation is an essential component of model evaluation as it enables large-scale testing and provides more stable and rapid feedback compared to purely human assessments. We introduce DreamEval, a comprehensive multimodal benchmark comprising four generation scenarios, containing 128 sub-tasks with 1,600 prompts. The scoring process is broken down into fine-grained visual-question-answer for each prompt, making evaluation more interpretable and deterministic. In particular, DreamEval also incorporates tiered difficulty levels that separately assess basic generation skills, advanced generation abilities, and higher-order understanding and reasoning capacity.

The overall results for instruction following are shown in Figure 1, and Figure~\ref{fig:multi_editing}, where the results are well aligned with human evaluations. Detailed performance across three levels of difficulty in the T2I and single-image editing tasks is shown in Figure~\ref{fig:multi_editing}. Several observations can be obtained: (1) Seedream 4.0 and GPT-4o outperform other models in instruction adherence, although Seedream 4.0 exhibits greater variability: it has a slightly lower average score, but its "best-of-4" results are better, suggesting that users can obtain better outputs through sampling. (2) Seedream 4.0 performs well at the Easy and Medium levels, demonstrating strong generative responsiveness; but its performance drops at the Hard level, especially in single-image editing. This highlights the need for improvement in multi-modal understanding and reasoning, which will be improved by scaling our models with related data.

\subsection{Inspire Creativity via Seedream 4.0}

We present several illustrative use cases of Seedream 4.0 in this section. These examples highlight only part of its potential, as further creative applications will emerge through user exploration.

\begin{figure*}[!t]
\centering
\includegraphics[width=\linewidth]{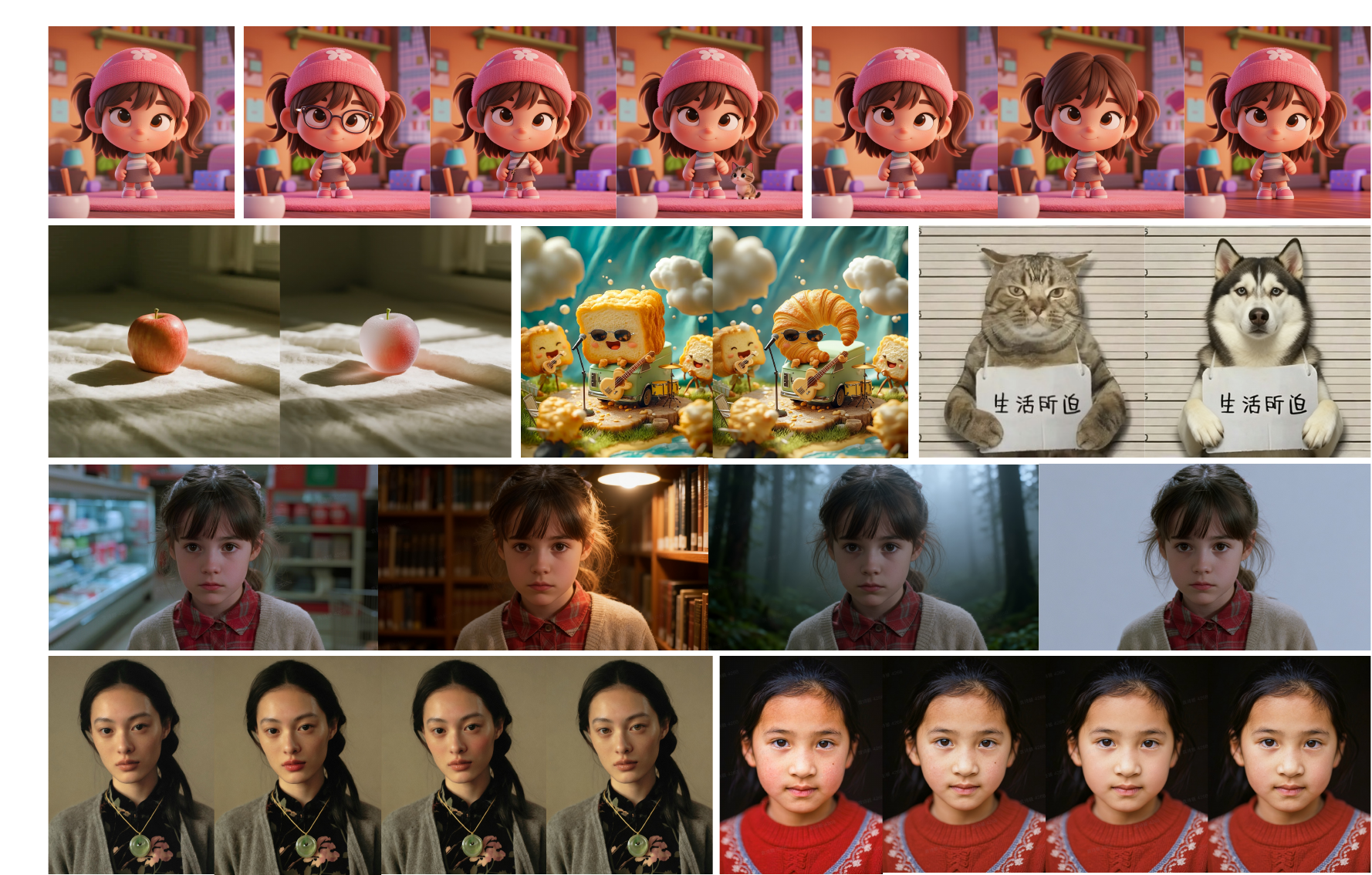}
\caption{Examples of precise editing.}
\label{fig:edit}
\end{figure*}

\subsubsection{Precise Editing}

Image editing has long been a critical challenge for generative models, with the main difficulty lying in achieving the desired modifications while preserving the majority of the original visual characteristics. Seedream 4.0 enables high-quality image editing solely from prompt-based input. It demonstrates strong instruction-following capability, delivering precise modifications while largely preserving the integrity of the surrounding visual content. As illustrated in Figure~\ref{fig:edit}, beyond canonical tasks such as addition, deletion, modification, and replacement, Seedream 4.0 shows remarkable performance in a variety of practical editing scenarios. For instance, in background replacement, it seamlessly integrates the foreground with other elements, and in portrait retouching, it delivers results that exhibit photographic realism.

\subsubsection{Flexible Reference}

\begin{figure*}[!t]
\centering
\includegraphics[width=\linewidth]{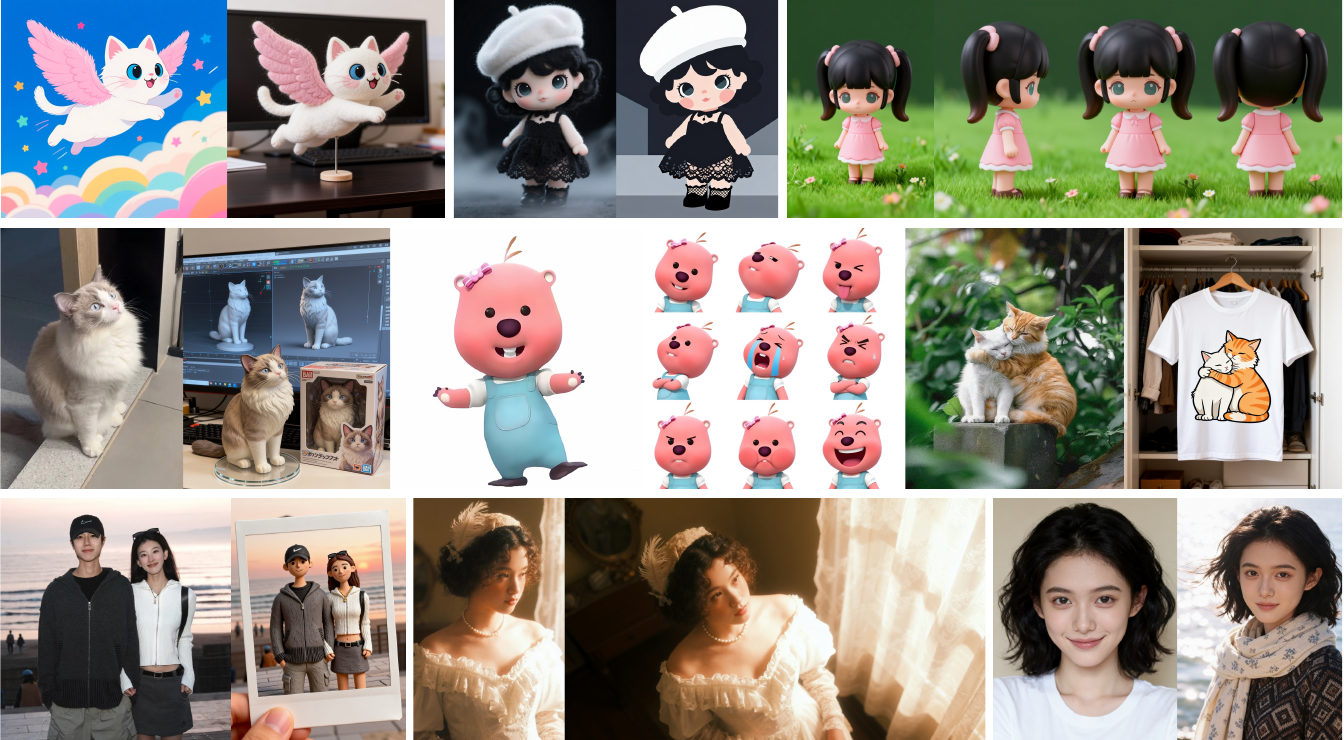}
\caption{Examples of reference generation.}
\label{fig:reference}
\end{figure*}

Unlike image editing, reference-based generation presents a more challenging trade-off between preservation and creativity. This difficulty arises from the inherently ambiguous definition of what should be preserved. It may be a person's ID or IP, a particular artistic style, or even an abstract concept. Consequently, the range of possible applications is broader and more diverse. As illustrated in Figure~\ref{fig:reference}, Seedream 4.0 supports seamless transformations across 2D/3D domains with varying viewpoints. It enables derivative designs such as dolls, clothes, or memes from a single reference image. In addition, benefiting from the strong consistency by Seedream 4.0, it can be effectively applied to identity-sensitive scenarios, including generating portrait photographs in different styles or creating characters for film.

\subsubsection{Visual Signal Controllable Generation}

\begin{figure*}[!t]
\centering
\includegraphics[width=\linewidth]{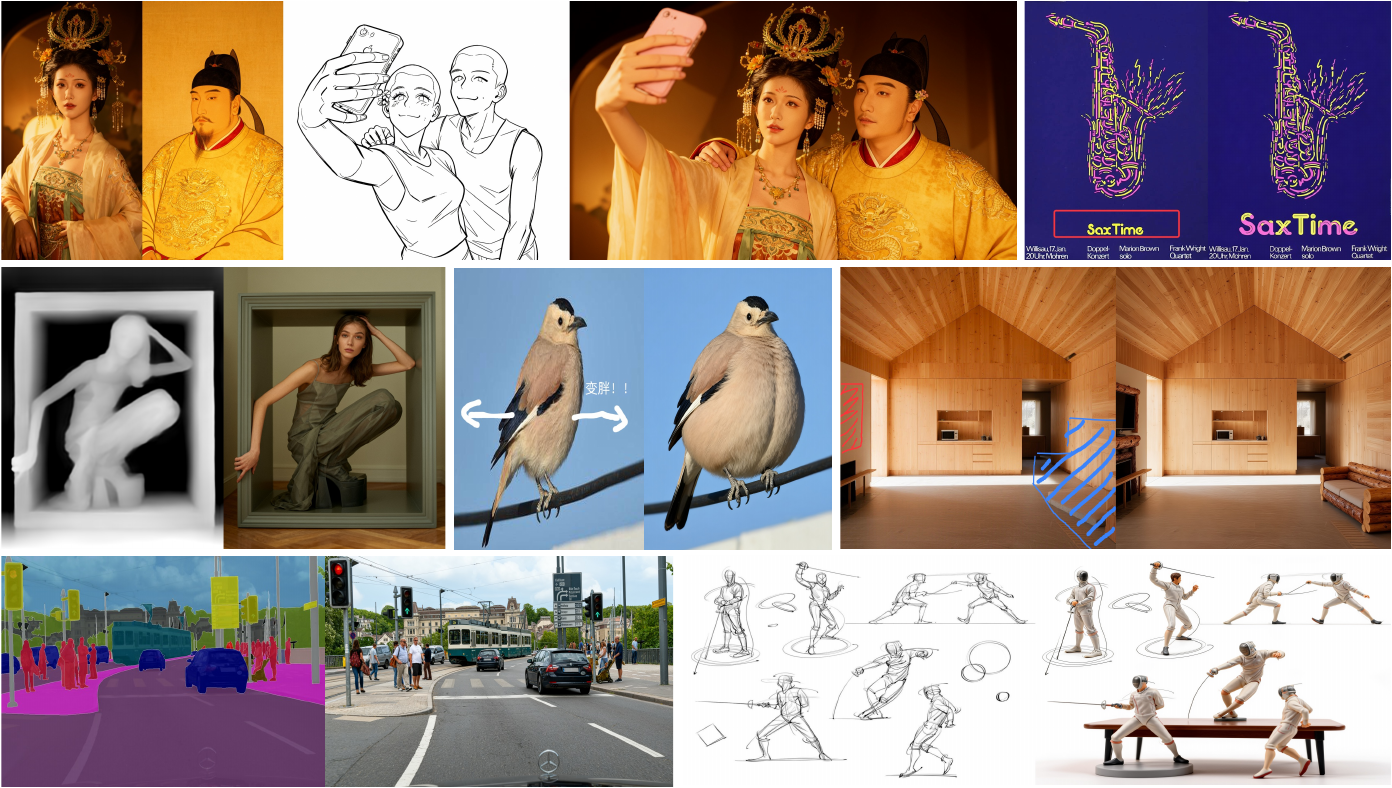}
\caption{Examples of visual signal controllable generation.}
\label{fig:visual_signal}
\end{figure*}

Visual signals, such as Canny edges, sketches, inpainting masks, or depth maps, have long been a crucial
component of controllable generation. They enable a transfer of information that is often difficult to describe through language, such as human poses or precise spatial layouts, thereby allowing for more accurate and targeted generation. Traditionally, this capability has been developed using multiple specialized models such as ControlNet \cite{zhang2023addingconditionalcontroltexttoimage,li2025controlnet}. In contrast, Seedream 4.0 natively integrates these functionalities with a single model. Beyond supporting the common forms of visual guidance, it also accommodates creative inputs through simple strokes or sketches, and even enables new multi-image compositions driven by visual signals. Illustrative examples are provided in Figure~\ref{fig:visual_signal}.

\subsubsection{In-Context Reasoning Generation}
\begin{figure*}[!t]
\centering
\includegraphics[width=\linewidth]{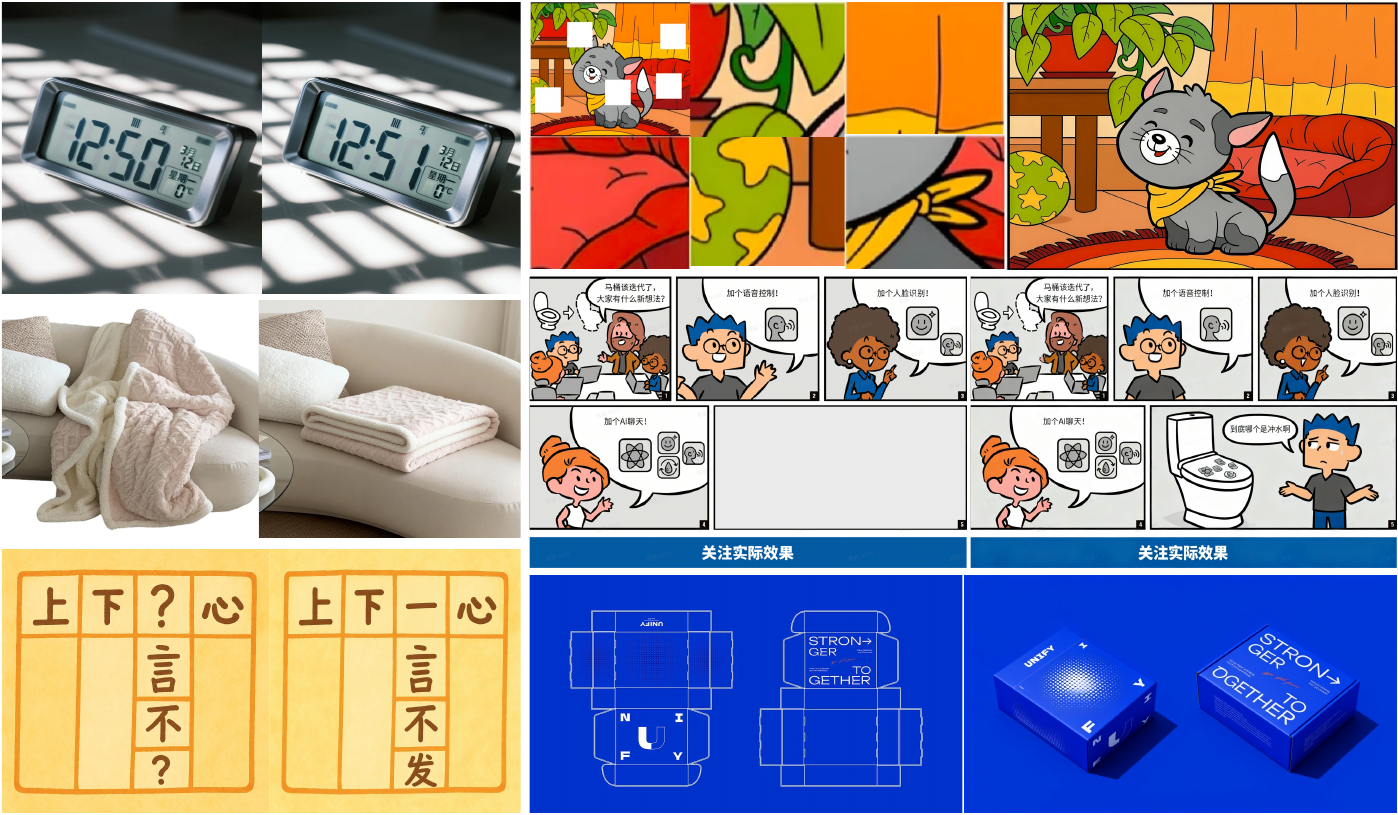}
\caption{Examples of reasoning generation.}
\label{fig:reasoning}
\end{figure*}

With the increasing intelligence of multimodal models, a new paradigm for in-context reasoning generation has emerged. Traditional image generation aims primarily at producing outputs that strictly follow the instructions given. In contrast, reasoning-based generation requires the model to go a step further: it must extract implicit contextual cues and infer plausible outcomes before genearating  the image. This process may involve expanding  the original prompt and interpreting reference images. As illustrated in Figure~\ref{fig:reasoning}, Seedream 4.0 demonstrates reasoning capabilities across  various  in-context understanding tasks, including interpreting  physical and temporal constraints of the real world, as well as imagining  three-dimensional space. Additionally , Seedream 4.0 can also perform tasks  such as puzzle solving, crossword filling, and comic continuation, all while faithfully preserving the visual style and fine-grained details of the given input.

\subsubsection{Multi-Image Reference Generation}

\begin{figure*}[!t]
\centering
\includegraphics[width=\linewidth]{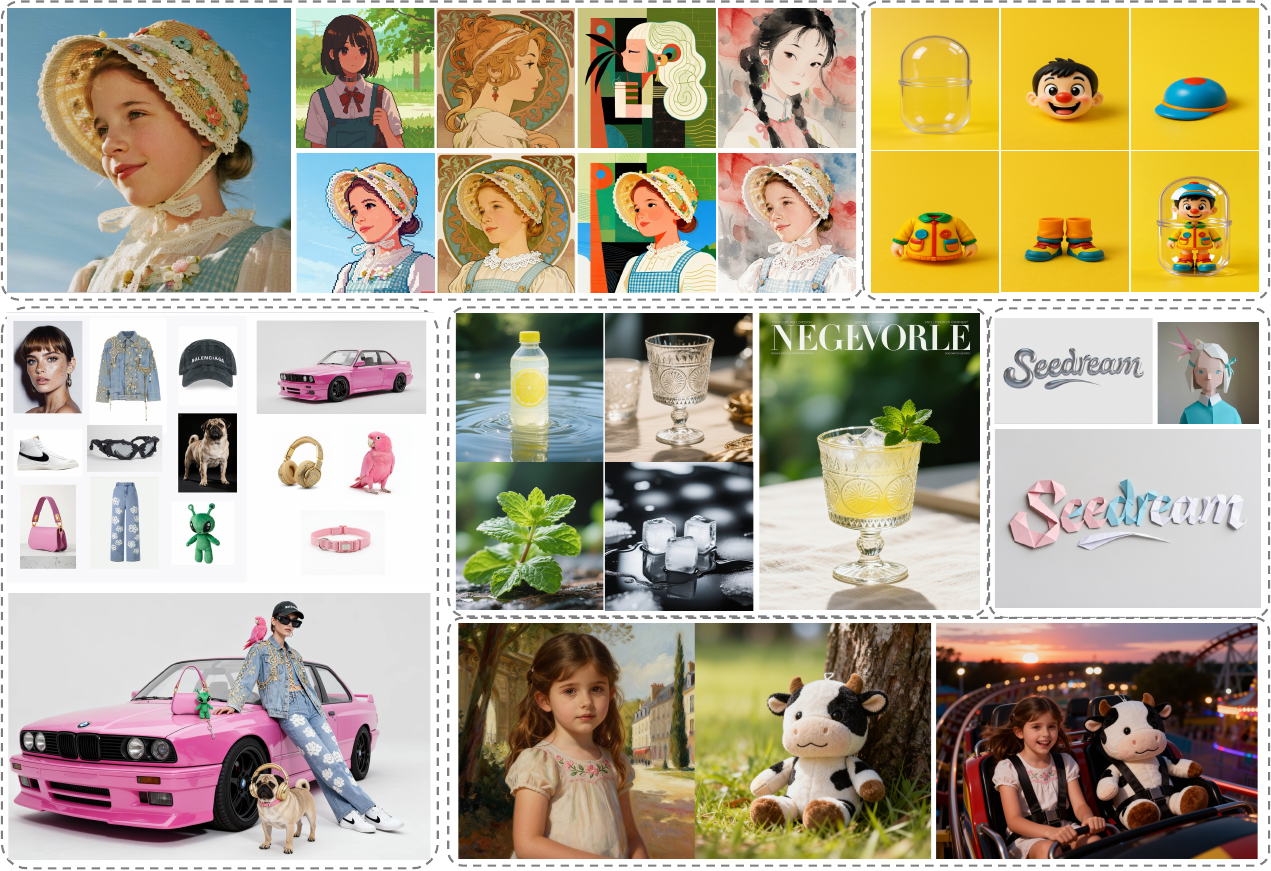}
\caption{Examples of multi-image composition.}
\label{fig:composition}
\end{figure*}

Benefiting from the richer information provided by multiple images, multi-image reference enables more imaginative and versatile applications. Beyond conventional tasks such as virtual try-on or image collage, it supports flexible composition of multiple characters or objects, as well as abstract style transfer. Unlike text conditioning, which requires explicitly specifying attributes or styles, multi-image editing compels the model to autonomously extract salient features from reference images and transfer them to the target. As illustrated in Figure~\ref{fig:composition}, Seedream 4.0 can handle reference-based editing with more than ten input images, while maintaining high fidelity in transferring abstract styles such as origami or Baroque aesthetics. Moreover, it effectively manages relative object scales and produces meaningful multi-object compositions, such as assembling mechanical parts, demonstrating a robust understanding of physical-world structures.

\subsubsection{Multi-Image Output}

\begin{figure*}[!t]
\centering
\includegraphics[width=\linewidth]{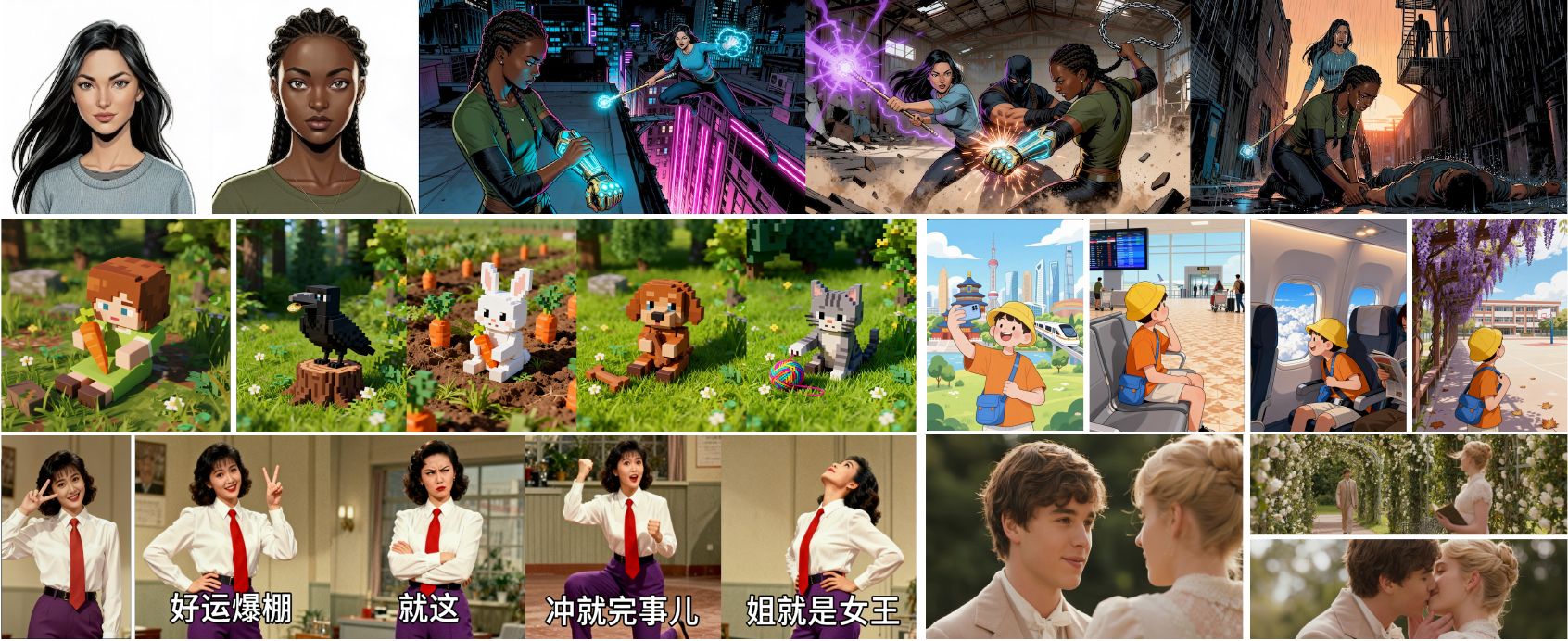}
\caption{Examples of multi-image output generation.}
\label{fig:multi_output}
\end{figure*}

Single-image generation is insufficient for many creative scenarios that require coherent multi-image outputs. Leveraging strong capabilities in global planning and in-context consistency, Seedream 4.0 supports the generation of image sequences that remain both character-consistent and stylistically aligned. As illustrated in Figure~\ref{fig:multi_output}, this enables sequential images generation based on given characters, which is particularly beneficial for storyboarding and comic creation. Seedream 4.0 can also produce sets of images with a consistent visual identity, a feature highly valuable for IP-based product design and the creation of emoji.

\subsubsection{Advanced Text Rendering}

\begin{figure*}[!t]
\centering
\includegraphics[width=\linewidth]{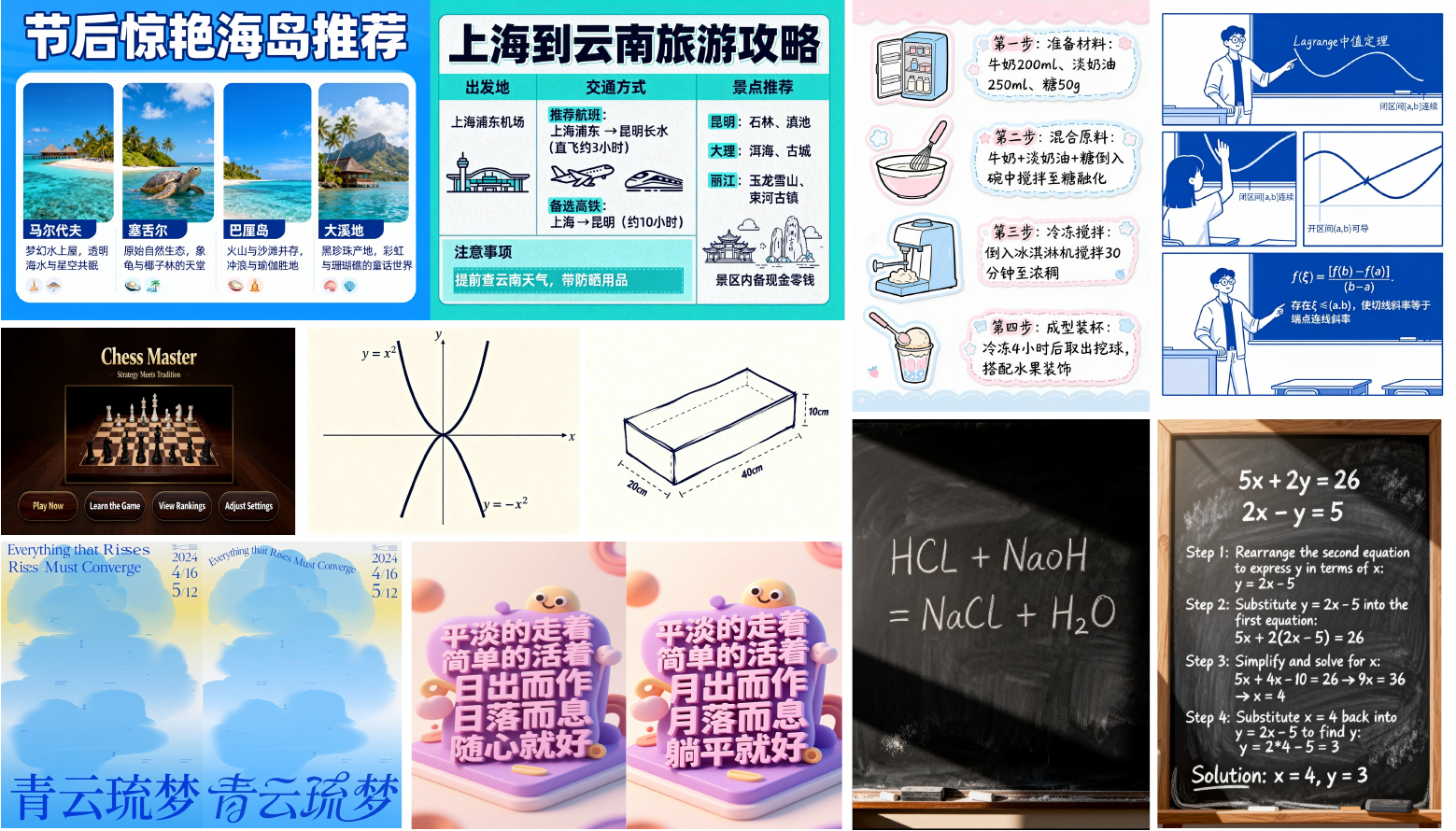}
\caption{Examples of advanced text rendering.}
\label{fig:text}
\end{figure*}

Seedream 4.0 introduces enhanced text-rendering capabilities that go beyond mere demonstration to serve practical applications. With intelligent understanding and extension as well as high-precision dense text rendering capabilities, it supports various complex text and graphic generation tasks, including designing layouts for user interfaces, posters, or schematics, as well as generating knowledge-intensive visualizations such as mathematical formulas, chemical equations, or statistical charts, as shown in Figure~\ref{fig:text}. Such capabilities make it feasible for the model to directly produce educational materials, technical manuals, or marketing content. In addition, Seedream 4.0 enables precise text-aware editing, including content replacement, layout adjustment, and font modification, thereby extending its rendering capacity to practical workflows and offering support for work-related scenarios.

\subsubsection{Adaptive Aspect Ratio and 4K Generation}

\begin{figure*}[!t]
\centering
\includegraphics[width=\linewidth]{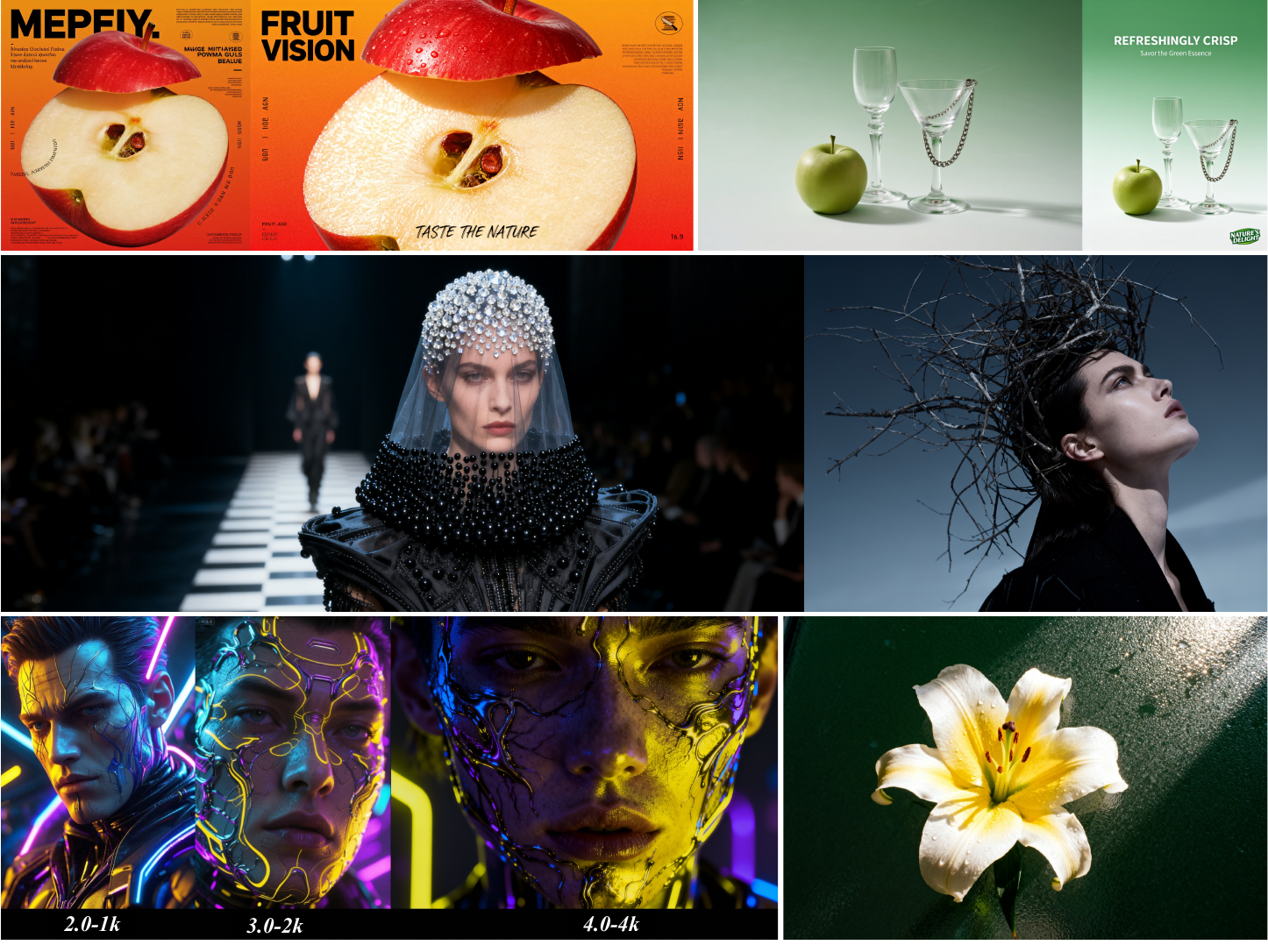}
\caption{Examples of adaptive ratio and 4K generation.}
\label{fig:4k}
\end{figure*}

Traditional generation models typically require a specified resolution, and selecting an unsuitable aspect ratio can lead to suboptimal composition and layout. Seedream 4.0 introduces an adaptive aspect ratio mechanism (while still supporting user-specified size), enabling the model to automatically adjust the canvas according to either the semantic requirements or the reference objects' shapes. As illustrated in Figure~\ref{fig:4k}, it allows the generation of images with more aesthetically pleasing and contextually appropriate compositions.
Moreover, Seedream 4.0 further extends its supporting resolution up to  4K. This advancement goes beyond research prototypes, delivering image quality suitable for commercial applications.

\section{Seedream 4.5}
\label{seedream4.5}
\begin{figure*}[!t]
    \centering
    \includegraphics[width=0.8\linewidth]{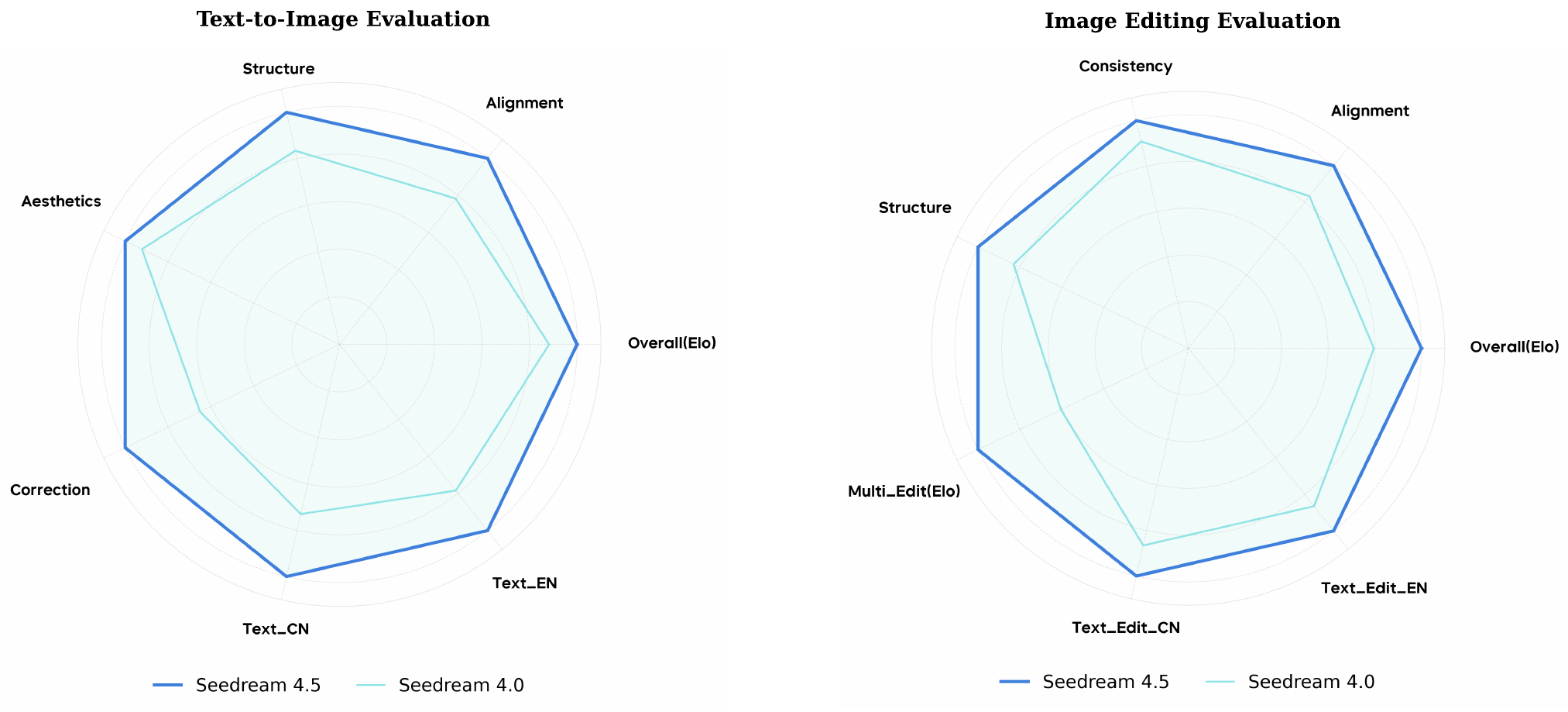}
    \caption{Comparisons between Seedream 4.5 and Seedream 4.0.}
    \label{fig:4l}
\end{figure*}

We further scale up both model size and training data, confirming the scalability of our framework and yielding a strengthened version, Seedream 4.5. It achieves significant improvements over Seedream 4.0 across all text-to-image and image editing tasks, including text–image alignment, structural fidelity, editing consistency, and fine-grained text rendering, as illustrated in Figure~\ref{fig:4l}. Notably, it exhibits markedly enhanced editing consistency, preserving identity-relevant features and fine-grained visual details from reference images, while accurately identifying and maintaining the target subject in multi-image composition scenarios. In addition, Seedream 4.5 provides stronger typographic rendering capabilities for dense text layouts such as posters and visual designs. These advances provide a better creative experience and superior output quality in Seedream 4.5.

\section{Conclusion}

In this report, we present Seedream 4.0, an advanced multimodal image generation framework that integrates an efficient and scalable diffusion transformer with a high-compression VAE, achieving more than ten times the acceleration of the previous Seedream 3.0 model, while delivering superior performance on all aspects. By performing joint post-training on T2I and image editing tasks, Seedream 4.0 provides strong multimodal generation capabilities that support diverse inputs and outputs. It demonstrates broad potential for creative exploration, including precise image editing, reference-based generation, multi-image composition, and multi-image output. 
In particular, the designed model architecture is highly efficient and stable, which allows us to scale it effectively, with considerable performance improvement achieved (in our on-going work).
Furthermore, with advanced inference acceleration technologies, Seedream 4.0 enables ultrafast image generation and editing at high resolutions. 
In addition, it also has the strong ability to support complicated content generation for professional scenarios, such as knowledge-centric generation, which is difficult to perform by previous models.
With its integration into platforms such as Doubao and Jimeng/Dreamina~\cite{dreamina}, Seedream 4.0 shows great potential to become a powerful productivity tool in content creation, for professional and everyday applications.

\clearpage

\bibliographystyle{plainnat}
\bibliography{main}

\begin{thebibliography}{26}
\providecommand{\natexlab}[1]{#1}
\providecommand{\url}[1]{\texttt{#1}}
\expandafter\ifx\csname urlstyle\endcsname\relax
  \providecommand{\doi}[1]{doi: #1}\else
  \providecommand{\doi}{doi: \begingroup \urlstyle{rm}\Url}\fi

\bibitem[artificialanalysis.ai(2025)]{aa_elo}
artificialanalysis.ai.
\newblock artificialanalysis.
\newblock https://artificialanalysis.ai/text-to-image/arena?tab=Leaderboard, 2025.

\bibitem[dreamina(2025)]{dreamina}
dreamina.
\newblock dreamina.
\newblock https://dreamina.capcut.com/, 2025.

\bibitem[Gao et~al.(2025)Gao, Gong, Guo, Hou, Lai, Li, Li, Lian, Liao, Liu, et~al.]{gao2025seedream}
Yu~Gao, Lixue Gong, Qiushan Guo, Xiaoxia Hou, Zhichao Lai, Fanshi Li, Liang Li, Xiaochen Lian, Chao Liao, Liyang Liu, et~al.
\newblock Seedream 3.0 technical report.
\newblock \emph{arXiv preprint arXiv:2504.11346}, 2025.

\bibitem[Gong et~al.(2025)Gong, Hou, Li, Li, Lian, Liu, Liu, Liu, Lu, Shi, et~al.]{gong2025seedream}
Lixue Gong, Xiaoxia Hou, Fanshi Li, Liang Li, Xiaochen Lian, Fei Liu, Liyang Liu, Wei Liu, Wei Lu, Yichun Shi, et~al.
\newblock Seedream 2.0: A native chinese-english bilingual image generation foundation model.
\newblock \emph{arXiv preprint arXiv:2503.07703}, 2025.

\bibitem[Google(2025)]{gemini2.5}
Google.
\newblock gemini2.5.
\newblock https://deepmind.google/models/gemini/image/, 2025.

\bibitem[Guo et~al.(2025)Guo, Wu, Zhu, Leng, Shi, Chen, Fan, Wang, Jiang, Wang, et~al.]{guo2025seed1}
Dong Guo, Faming Wu, Feida Zhu, Fuxing Leng, Guang Shi, Haobin Chen, Haoqi Fan, Jian Wang, Jianyu Jiang, Jiawei Wang, et~al.
\newblock Seed1. 5-vl technical report.
\newblock \emph{arXiv preprint arXiv:2505.07062}, 2025.

\bibitem[Labs(2023)]{flux2023}
Black~Forest Labs.
\newblock Flux.
\newblock \url{https://github.com/black-forest-labs/flux}, 2023.

\bibitem[Labs et~al.(2025)Labs, Batifol, Blattmann, Boesel, Consul, Diagne, Dockhorn, English, English, Esser, Kulal, Lacey, Levi, Li, Lorenz, Müller, Podell, Rombach, Saini, Sauer, and Smith]{labs2025flux1kontextflowmatching}
Black~Forest Labs, Stephen Batifol, Andreas Blattmann, Frederic Boesel, Saksham Consul, Cyril Diagne, Tim Dockhorn, Jack English, Zion English, Patrick Esser, Sumith Kulal, Kyle Lacey, Yam Levi, Cheng Li, Dominik Lorenz, Jonas Müller, Dustin Podell, Robin Rombach, Harry Saini, Axel Sauer, and Luke Smith.
\newblock Flux.1 kontext: Flow matching for in-context image generation and editing in latent space, 2025.
\newblock URL \url{https://arxiv.org/abs/2506.15742}.

\bibitem[Li et~al.(2025)Li, Yang, Kuang, Wu, Wang, Xiao, and Chen]{li2025controlnet}
Ming Li, Taojiannan Yang, Huafeng Kuang, Jie Wu, Zhaoning Wang, Xuefeng Xiao, and Chen Chen.
\newblock Controlnet++: Improving conditional controls with efficient consistency feedback.
\newblock In \emph{European Conference on Computer Vision}, pages 129--147. Springer, 2025.

\bibitem[Lin et~al.(2025)Lin, Xia, Ren, Yang, Xiao, and Jiang]{lin2025diffusion}
Shanchuan Lin, Xin Xia, Yuxi Ren, Ceyuan Yang, Xuefeng Xiao, and Lu~Jiang.
\newblock Diffusion adversarial post-training for one-step video generation.
\newblock \emph{arXiv preprint arXiv:2501.08316}, 2025.

\bibitem[Liu et~al.(2025)Liu, Liu, Liang, Li, Liu, Wang, Wan, Zhang, and Ouyang]{liu2025flow}
Jie Liu, Gongye Liu, Jiajun Liang, Yangguang Li, Jiaheng Liu, Xintao Wang, Pengfei Wan, Di~Zhang, and Wanli Ouyang.
\newblock Flow-grpo: Training flow matching models via online rl.
\newblock \emph{arXiv preprint arXiv:2505.05470}, 2025.

\bibitem[Lou et~al.(2025)Lou, Sun, Liang, Qu, Shen, Wang, Li, Yang, and Wu]{lou2025adacot}
Chenwei Lou, Zewei Sun, Xinnian Liang, Meng Qu, Wei Shen, Wenqi Wang, Yuntao Li, Qingping Yang, and Shuangzhi Wu.
\newblock Adacot: Pareto-optimal adaptive chain-of-thought triggering via reinforcement learning.
\newblock \emph{arXiv preprint arXiv:2505.11896}, 2025.

\bibitem[Lu et~al.(2025{\natexlab{a}})Lu, Ren, Xia, Lin, Wang, Xiao, Ma, Xie, and Lai]{lu2025adversarial}
Yanzuo Lu, Yuxi Ren, Xin Xia, Shanchuan Lin, Xing Wang, Xuefeng Xiao, Andy~J Ma, Xiaohua Xie, and Jian-Huang Lai.
\newblock Adversarial distribution matching for diffusion distillation towards efficient image and video synthesis.
\newblock \emph{arXiv preprint arXiv:2507.18569}, 2025{\natexlab{a}}.

\bibitem[Lu et~al.(2025{\natexlab{b}})Lu, Xia, Zhang, Kuang, Zheng, Ren, and Xiao]{lu2025hyperbagelunifiedaccelerationframework}
Yanzuo Lu, Xin Xia, Manlin Zhang, Huafeng Kuang, Jianbin Zheng, Yuxi Ren, and Xuefeng Xiao.
\newblock Hyper-bagel: A unified acceleration framework for multimodal understanding and generation, 2025{\natexlab{b}}.
\newblock URL \url{https://arxiv.org/abs/2509.18824}.

\bibitem[OpenAI(2025)]{gpt-4o}
OpenAI.
\newblock Gpt-4o.
\newblock \url{https://openai.com/index/introducing-4o-image-generation/}, 2025.

\bibitem[OpenAI et~al.(2024)OpenAI, :, Hurst, and et~al.]{openai2024gpt4ocard}
OpenAI, :, Aaron Hurst, and Adam~Lerer et~al.
\newblock Gpt-4o system card, 2024.
\newblock URL \url{https://arxiv.org/abs/2410.21276}.

\bibitem[Ren et~al.(2025)Ren, Xia, Lu, Zhang, Wu, Xie, Wang, and Xiao]{ren2025hyper}
Yuxi Ren, Xin Xia, Yanzuo Lu, Jiacheng Zhang, Jie Wu, Pan Xie, Xing Wang, and Xuefeng Xiao.
\newblock Hyper-sd: Trajectory segmented consistency model for efficient image synthesis.
\newblock \emph{Advances in Neural Information Processing Systems}, 37:\penalty0 117340--117362, 2025.

\bibitem[Rombach et~al.(2022)Rombach, Blattmann, Lorenz, Esser, and Ommer]{rombach2022high}
Robin Rombach, Andreas Blattmann, Dominik Lorenz, Patrick Esser, and Bj{\"o}rn Ommer.
\newblock High-resolution image synthesis with latent diffusion models.
\newblock In \emph{CVPR}, pages 10684--10695, 2022.

\bibitem[Shao et~al.(2025)Shao, Xia, Yang, Ren, Wang, and Xiao]{shao2025rayflow}
Huiyang Shao, Xin Xia, Yuhong Yang, Yuxi Ren, Xing Wang, and Xuefeng Xiao.
\newblock Rayflow: Instance-aware diffusion acceleration via adaptive flow trajectories.
\newblock \emph{arXiv preprint arXiv:2503.07699}, 2025.

\bibitem[Shi et~al.(2024{\natexlab{a}})Shi, Wang, and Huang]{seededit2024}
Yichun Shi, Peng Wang, and Weilin Huang.
\newblock Seededit: Align image re-generation to image editing, 2024{\natexlab{a}}.
\newblock URL \url{https://arxiv.org/abs/2411.06686}.

\bibitem[Shi et~al.(2024{\natexlab{b}})Shi, Wang, and Huang]{shi2024seededit}
Yichun Shi, Peng Wang, and Weilin Huang.
\newblock Seededit: Align image re-generation to image editing.
\newblock \emph{arXiv preprint arXiv:2411.06686}, 2024{\natexlab{b}}.

\bibitem[Wu et~al.(2025{\natexlab{a}})Wu, Li, Zhou, Lin, Gao, Yan, ming Yin, Bai, Xu, Chen, Chen, Tang, Zhang, Wang, Yang, Yu, Cheng, Liu, Li, Zhang, Meng, Wei, Ni, Chen, Cao, Peng, Qu, Wu, Wang, Yu, Wen, Feng, Xu, Wang, Zhang, Zhu, Wu, Cai, and Liu]{wu2025qwenimagetechnicalreport}
Chenfei Wu, Jiahao Li, Jingren Zhou, Junyang Lin, Kaiyuan Gao, Kun Yan, Sheng ming Yin, Shuai Bai, Xiao Xu, Yilei Chen, Yuxiang Chen, Zecheng Tang, Zekai Zhang, Zhengyi Wang, An~Yang, Bowen Yu, Chen Cheng, Dayiheng Liu, Deqing Li, Hang Zhang, Hao Meng, Hu~Wei, Jingyuan Ni, Kai Chen, Kuan Cao, Liang Peng, Lin Qu, Minggang Wu, Peng Wang, Shuting Yu, Tingkun Wen, Wensen Feng, Xiaoxiao Xu, Yi~Wang, Yichang Zhang, Yongqiang Zhu, Yujia Wu, Yuxuan Cai, and Zenan Liu.
\newblock Qwen-image technical report, 2025{\natexlab{a}}.
\newblock URL \url{https://arxiv.org/abs/2508.02324}.

\bibitem[Wu et~al.(2025{\natexlab{b}})Wu, Gao, Ye, Li, Li, Guo, Liu, Xue, Hou, Liu, et~al.]{wu2025rewarddance}
Jie Wu, Yu~Gao, Zilyu Ye, Ming Li, Liang Li, Hanzhong Guo, Jie Liu, Zeyue Xue, Xiaoxia Hou, Wei Liu, et~al.
\newblock Rewarddance: Reward scaling in visual generation.
\newblock \emph{arXiv preprint arXiv:2509.08826}, 2025{\natexlab{b}}.

\bibitem[Xu et~al.(2024)Xu, Liu, Wu, Tong, Li, Ding, Tang, and Dong]{xu2024imagereward}
Jiazheng Xu, Xiao Liu, Yuchen Wu, Yuxuan Tong, Qinkai Li, Ming Ding, Jie Tang, and Yuxiao Dong.
\newblock Imagereward: Learning and evaluating human preferences for text-to-image generation.
\newblock \emph{Advances in Neural Information Processing Systems}, 36, 2024.

\bibitem[Xue et~al.(2025)Xue, Wu, Gao, Kong, Zhu, Chen, Liu, Liu, Guo, Huang, et~al.]{xue2025dancegrpo}
Zeyue Xue, Jie Wu, Yu~Gao, Fangyuan Kong, Lingting Zhu, Mengzhao Chen, Zhiheng Liu, Wei Liu, Qiushan Guo, Weilin Huang, et~al.
\newblock Dancegrpo: Unleashing grpo on visual generation.
\newblock \emph{arXiv preprint arXiv:2505.07818}, 2025.

\bibitem[Zhang et~al.(2023)Zhang, Rao, and Agrawala]{zhang2023addingconditionalcontroltexttoimage}
Lvmin Zhang, Anyi Rao, and Maneesh Agrawala.
\newblock Adding conditional control to text-to-image diffusion models, 2023.
\newblock URL \url{https://arxiv.org/abs/2302.05543}.

\end{thebibliography}

\clearpage

\beginappendix
\section{Contributions and Acknowledgments}
\label{contributions}

All contributors of Seedream are listed in alphabetical order by their last names.

\subsection{Core Contributors}
Yunpeng Chen,
Yu Gao,
Lixue Gong,
Meng Guo,
Qiushan Guo,
Zhiyao Guo,
Xiaoxia Hou,
Weilin Huang,
Yixuan Huang,
Xiaowen Jian,
Huafeng Kuang,
Zhichao Lai,
Fanshi Li,
Liang Li,
Xiaochen Lian,
Chao Liao,
Liyang Liu,
Wei Liu,
Yanzuo Lu,
Zhengxiong Luo,
Tongtong Ou,
Guang Shi,
Yichun Shi,
Shiqi Sun,
Yu Tian,
Zhi Tian,
Peng Wang,
Rui Wang,
Xun Wang,
Ye Wang,
Guofeng Wu,
Jie Wu,
Wenxu Wu,
Yonghui Wu,
Xin Xia,
Xuefeng Xiao,
Shuang Xu,
Xin Yan,
Ceyuan Yang,
Jianchao Yang,
Zhonghua Zhai,
Chenlin Zhang,
Heng Zhang,
Qi Zhang,
Xinyu Zhang,
Yuwei Zhang,
Shijia Zhao,
Wenliang Zhao,
Wenjia Zhu

\subsection{Contributors}
Haoshen Chen,
Kaixi Chen,
Tiantian Cheng,
Fei Ding,
Xiaojing Dong,
Xin Dong,
Yiming Fan,
Yongde Ge,
Shucheng Guo,
Bibo He,
Jiaao He,
Zhuo Jiang,
Lurui Jin,
Hongwei Kou,
Bo Li,
Changchun Li,
Hao Li,
Huixia Li,
Jiashi Li,
Yameng Li,
Ying Li,
Yiying Li,
Zijie Li,
Heng Lin,
Zhijie Lin,
Gaohong Liu,
Mingcong Liu,
Shu Liu,
Zuxi Liu,
Zhangfan Lu,
Xiaonan Nie,
Shuang Ouyang,
Ronggui Peng,
Keer Qin,
Xudong Sun,
Yang Tai,
Rupeng Tian,
Lei Wang,
Sen Wang,
Xuanda Wang,
Yinuo Wang,
Shaojin Wu,
Xiaohu Wu,
Wenpeng Xiao,
Yihang Yang,
Yao Yao,
Linxiao Yuan,
Dingyun Zhang,
Kai Zhang,
Manlin Zhang,
Tao Zhang,
Xinlei Zhang,
Yanling Zhang,
Yun Zhang,
Zixuan Zhang,
Fengxuan Zhao,
Hao Zheng,
Jianbin Zheng














\end{CJK*}
\end{document}